\documentclass[letterpaper]{article}
\usepackage[english]{babel}
\usepackage{times}
\usepackage{graphicx} 
\usepackage{subcaption}
\usepackage[round,sort]{natbib}
\usepackage{algorithm}
\usepackage{algorithmic}
\usepackage{hyperref}

\usepackage[accepted]{icml2017}
\usepackage[T1]{fontenc}
\usepackage[utf8x]{inputenc}

\icmltitlerunning{Inspiring Computer Vision System Solutions (ICVSS)}
\begin{document}
\twocolumn[
\icmltitle{Inspiring Computer Vision System Solutions}

\begin{icmlauthorlist}
\icmlauthor{Julian Zilly}{1}
\icmlauthor{Amit Boyarski}{2}
\icmlauthor{Micael Carvalho}{3}
\icmlauthor{Amir Atapour Abarghouei}{4}
\icmlauthor{Konstantinos Amplianitis}{5}
\icmlauthor{Aleksandr Krasnov}{6}
\icmlauthor{Massimiliano Mancini}{7}
\icmlauthor{Hern\'an Gonzalez}{8}
\icmlauthor{Riccardo Spezialetti}{9}
\icmlauthor{Carlos Sampedro P\'erez}{10}
\icmlauthor{Hao Li}{11}
\end{icmlauthorlist}

\icmlaffiliation{1}{ETH Z\"urich, Switzerland}
\icmlaffiliation{2}{Technion, Haifa, Israel}
\icmlaffiliation{3}{Universit\'e Pierre et Marie Curie, LIP6, France}
\icmlaffiliation{4}{Durham University, United Kingdom}
\icmlaffiliation{5}{Trinity College Dublin, Ireland}
\icmlaffiliation{6}{Apple Inc., USA}
\icmlaffiliation{7}{Sapienza University of Rome, Italy}
\icmlaffiliation{8}{Paris-Sud University, France}
\icmlaffiliation{9}{University of Bologna, Italy}
\icmlaffiliation{10}{Centre for Automation and Robotics (CAR) CSIC-UPM, Madrid, Spain}
\icmlaffiliation{11}{University of Southern California, Department of Computer Science, USA \& Pinscreen, Inc., USA}

\icmlcorrespondingauthor{Julian Zilly}{jzilly@ethz.ch}
\icmlkeywords{Computer Vision, Productive Failure}

\vskip 0.3in
]
\printAffiliationsAndNotice{}

\begin{abstract}
\textit{The "digital Michelangelo project" was a seminal computer vision project in the early 2000's that pushed the capabilities of acquisition systems and involved multiple people from diverse fields, many of whom are now leaders in industry and academia. Reviewing this project with modern eyes provides us with the opportunity to reflect on several issues, relevant now as then to the field of computer vision and research in general, that go beyond the technical aspects of the work.}

This article was written in the context of a reading group competition at the week-long International Computer Vision Summer School 2017 (ICVSS) on Sicily, Italy. To deepen the participants understanding of computer vision and to foster a sense of community, various reading groups were tasked to highlight important lessons which may be learned from provided literature, going beyond the contents of the paper. This report is the winning entry of this guided discourse (Fig. \ref{fig:reading_group}). The authors closely examined the origins, fruits and most importantly lessons about research in general which may be distilled from the "digital Michelangelo project". Discussions leading to this report were held within the group as well as with Hao Li, the group mentor.
\end{abstract}

\section{Introduction}

\begin{figure}
\includegraphics[width=.95\linewidth]{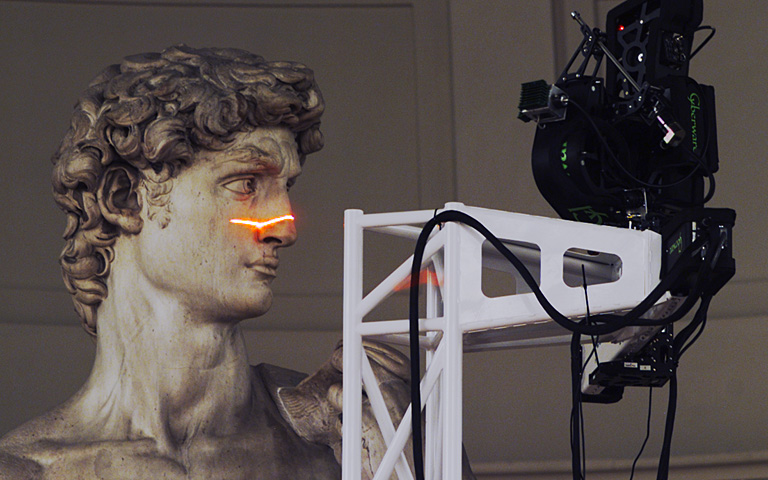}
\caption{Photograph of the scanning of the head of Michelangelo's "David" statue photographed by Marc Levoy and Paul Debevec, March 28, 1999.}
\label{fig:david_scanning}
\end{figure}

The seminal paper "The digital Michelangelo project" \cite{digital_michelangelo} encapsulates the design and implementation of software and hardware requirements for the digitization of large historical artifacts, mostly including Michelangelo’s sculpture artwork. The project involved a team of 30 people who spent a year scanning 10 statues created by Michelangelo, two building interiors and a marble map of ancient Rome, using a state-of-the-art of the time laser scanner mounted on a customized gantry. A sample photograph of this process is shown in Fig. \ref{fig:david_scanning}.

The project was conducted in the wild (non-laboratory conditions) and extreme precision was required to capture tiny bumps, and chisel marks intentionally left by the artist. Furthermore, due to the fragility of the subjects and their cultural and historical value, a non-contact mode of digitization had to be utilized to avoid harming the artwork. All these factors contributed to the design of the scanner.

The paper contains an exhaustive description and analysis of the challenges encountered during the project, including, but not limited to, optical characteristics of marble in general, the bias caused by subsurface scattering in the statues with respect to the type of marble and their coating, poor color fidelity due to the cross-talk between the laser and the broadband luminaire, configuration of the gantry and the scan head, deflection of the gantry during panning and tilting, alignment of the scans, and other software and hardware issues alike.

Calibration was performed in six pre-planned stages to ensure accuracy and repeatability of the system. During the post-processing step, range scans had to be manually aligned by the users, which in the end proved to be very time-consuming. Acquired color information was mapped onto the mesh after correcting for the ambient illumination, distortions, chromatic aberration, and radiometric effects. Every color associated with a mesh vertex was converted to reflectance through deshading. Then, when different color images were captured for a vertex, the reflectance would blend following a series of predefined rules. Additionally, due to the system's restrictions with respect to time, special measures were put in place to handle the large amount of collected data (two billion polygons and 36 gigabytes for the statue of David alone). Incredible care in planning and preparation helped in the realization of the project, and the encountered challenges have paved the way to new directions in research.

\section{Computer vision - a journey from past to future}
With the advent of deep learning, it is easy to forget that algorithms tell only half the story of computer vision. The other half is told by sensors, which provide a bridge between the real and the digital world. 
Indeed, the term “machine vision” refers both to the ability of a machine to assign meaning to an image, as well as to its ability to acquire this image. 
 
This versatility of machine vision systems is manifested in both new modalities captured such as hyperspectral and light field imaging \cite{chang2003hyperspectral, ng2005light}, and in new active and passive acquisition methods such as LIDAR \cite{lidar} and Dynamic Vision Sensors \cite{dynamic_vision_sensor}. The 3D scanning system described in \cite{digital_michelangelo} is an example of such a system. In years past, it has inspired an astonishing number of engineering efforts to gain more precise sensors for vision tasks.

Works like \cite{digital_michelangelo} can be considered part of the classical computer vision canon, relying heavily on the classical sciences. One could ask what the role of geometry and physics is in the emerging age of data-based computer vision. Apart from the apparent need for acquisition systems to provide high quality datasets, it is important to remember that developing such systems provides us not only with the mechanism to acquire high quality data, but also with a better understanding of the physical world which in turn can be used to simulate such data. 

For decades, computer vision has followed a bottom-up approach, trying to devise models that capture exact interactions between vision systems and the physical world. Despite the inability of such models to provide state-of-the-art results on par with human capabilities, they can be leveraged to support newer techniques through their integration in machine learning pipelines. This leads us to the encouraging thought that many of the models developed by the previous generation of researchers will not fall into oblivion.

\section{Roots of the paper}

In science we often remind ourselves that we are standing on the shoulders of giants. With this in mind, we would like to shed some light on the sources, the influences that inspired “The digital Michelangelo project” \cite{digital_michelangelo}, around which this report is centered. 
If we make the analogy that the mentioned paper is the very trunk of a tree, this report aims to uncover both the roots which made the trunk possible as well as the leaves which have originated from the trunk, respectively. 

\begin{figure*}
\centering
\begin{subfigure}{.5\textwidth}
  \centering
  \includegraphics[width=.97\linewidth]{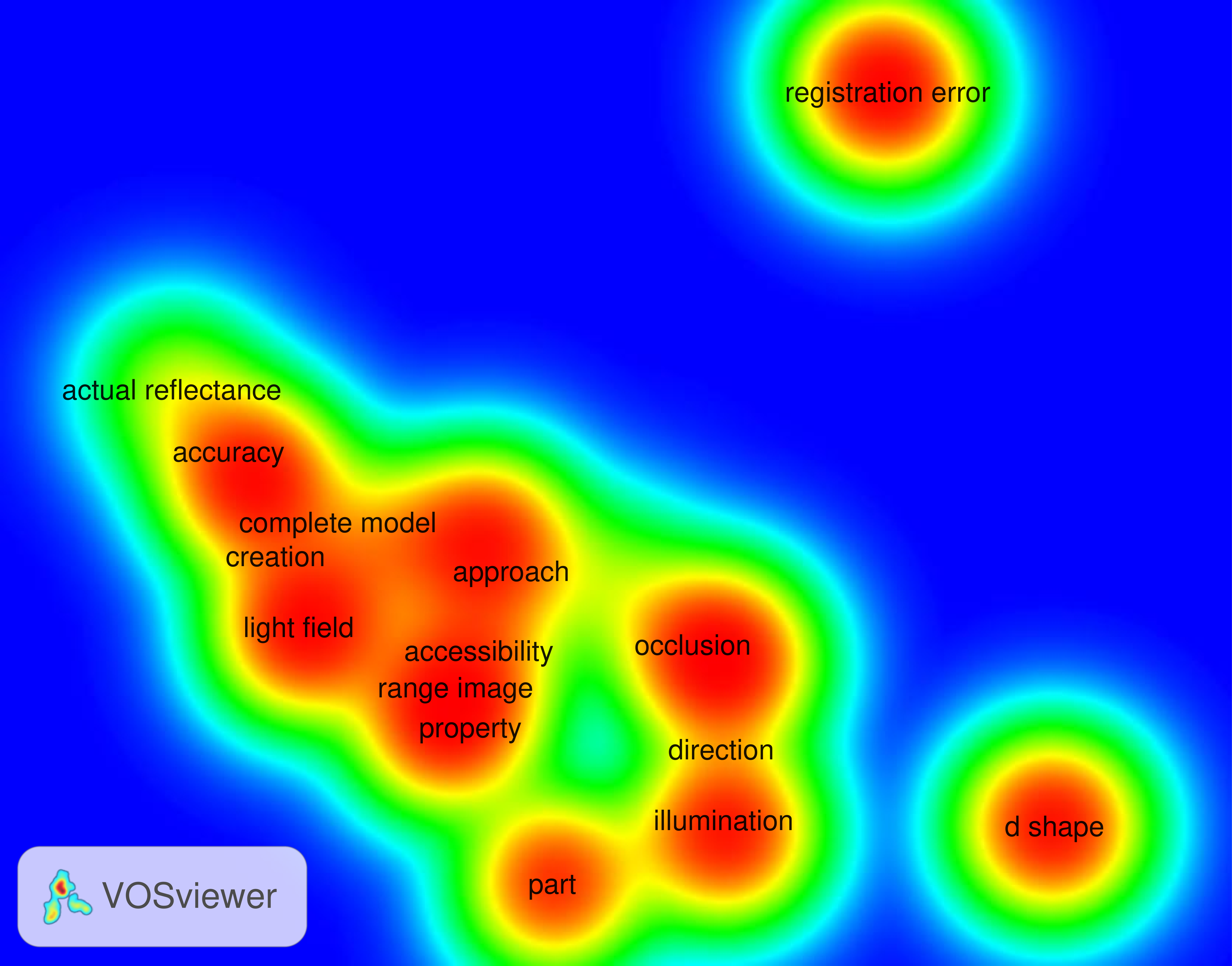}
  \caption{}
  \label{fig:fig1}
\end{subfigure}%
\begin{subfigure}{.5\textwidth}
  \centering
  \includegraphics[width=.97\linewidth]{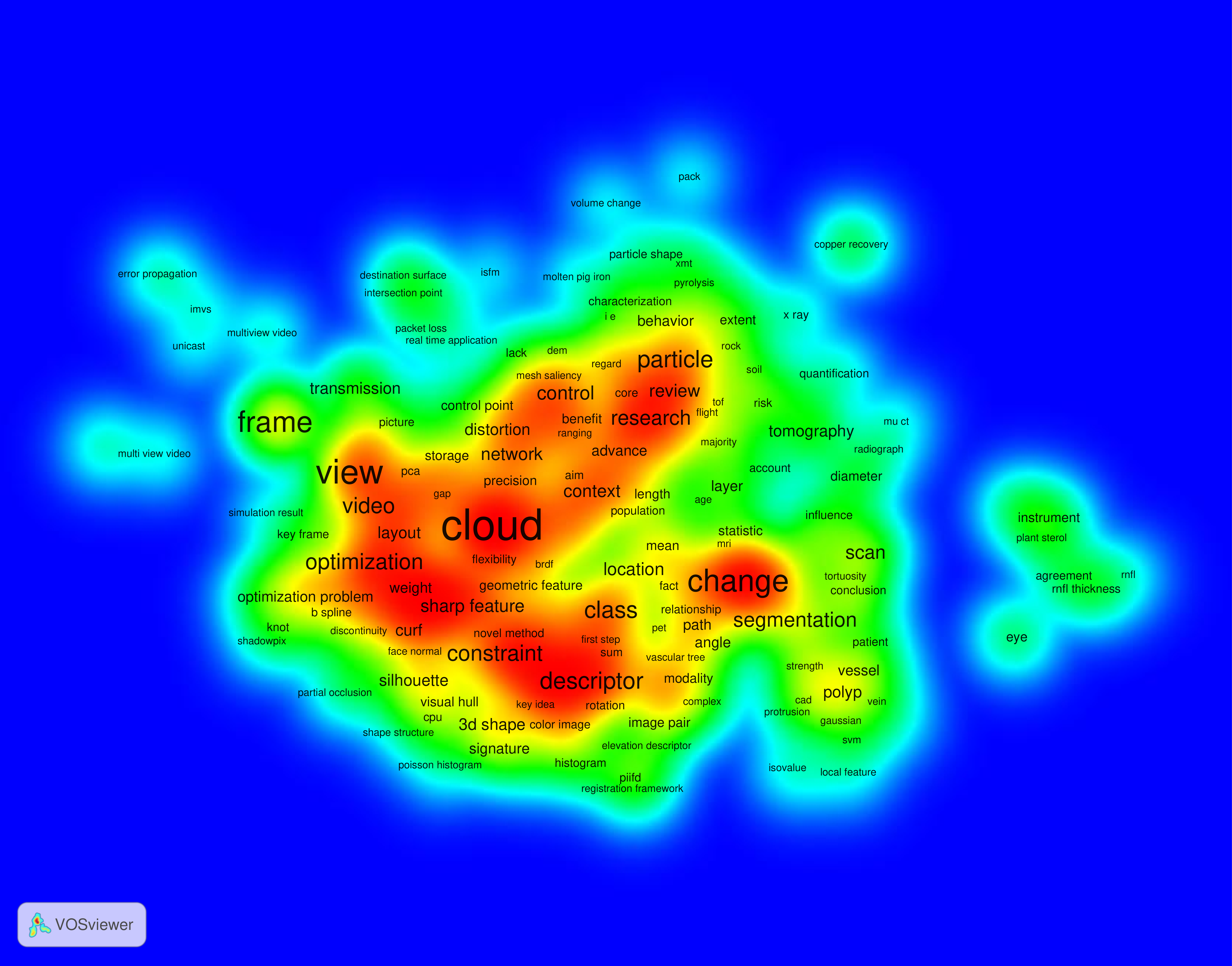}
  \caption{}
  \label{fig:fig2}
\end{subfigure}
\centering
\caption{Visualization of abstract themes present in the roots and leaves of provided paper. Subfigure a) showcases the main themes of the papers cited in the seminal work \cite{digital_michelangelo} which this report starts out with. Subfigure b) highlights the abstract themes and titles of most of the papers citing the mentioned seminal work. Illustrations were created using VOSViewer \cite{vosviewer}. \textbf{(Best viewed in color)}}
\label{fig:fig}
\end{figure*}

\textbf{Roots:} The ideas that have most influenced the digital Michelangelo project work \cite{digital_michelangelo} cover solutions and approaches to various problems that naturally arose in the course of the project. A visualization of the main ideas that influenced the digital Michelangelo project paper is depicted in Fig. \ref{fig:fig1} based on the literature cited in the paper. 

Starting out with scanning large marble statues, object models had to be created from multiple views \cite{object_modeling}. To ensure a realistic rendering later on, high quality textures had to be synthesized \cite{texture_synthesis}. Likewise, new techniques were developed to register data as points and work with these new data structures \cite{surface_registration, point_large_meshes}. Other interesting hurdles to overcome were dealing with large amounts of data \cite{multiview_large_data} and planning how to acquire data intelligently \cite{occlusions, next_view}.

Additionally, very specific real-world challenges had to be overcome such as the modeling of weathered stone \cite{weathered_stone} and camera calibration \cite{camera_calibration}. 

As we will lay out in the remainder of this report, many of these problems and their initial solutions provided a fertile ground for exciting new research directions.

\section{Leaves of the paper}

Branching off from the digital Michelangelo project \cite{digital_michelangelo}, a variety of interesting research directions emerged. 

\textbf{Point-based computer graphics:} As one of the early works, the digital Michelangelo project pioneered the usage of point clouds to align acquired scans. Since then, an active research community has developed that has improved upon many of the original techniques such as a more recent work on 3D point-based image fusion \cite{point_based_fusion}.

\textbf{Large scale data acquisition and processing:}
Acquiring high-quality data oftentimes sets the stage to overcome difficult challenges in computer vision and other disciplines. In this spirit, they clearly demonstrated the ability to both acquire and compute a large amount of data. Similarly, the published data of the scanned statues enabled many future works. This has set a positive example for later datasets published in the field, e.g. \cite{new_big_dataset}.

\textbf{Demonstrating the feasibility of difficult vision projects:}
Lastly, we would like to make the point that by overcoming the many challenges associated with such a demanding project, similar moon-shot endeavors are encouraged. 
Likewise, the challenging task they attacked in the year 2000 has since been partially re-addressed with more advanced methods and/or less specialized sensors as shown in \cite{same_but_with_camera}.

Figure \ref{fig:fig2} illustrates the main ideas of works citing the discussed digital Michelangelo project. It is interesting to note that the keywords "Cloud", "Descriptor" and "Tomography" appear prominently in these works. This serves to illustrate the previous argument that many different types of works have a connection and are at least partially made possible by the discussed paper.

Finally, there is also a lot of engineering value to this one-of-a-kind project, overcoming issues unique to the domain and precision level of its task. Particularly laudable were the proper documentation of solutions at \textbf{detail level which allows for result reproduction} and the upfront mentioning of difficulties and issues. 

\section{Discussion}

Taking the digital Michelangelo project as a successful example of a large-scale computer vision project, we would like to extract some key insights that can potentially aid us in conducting similarly successful large-scale projects in the future.

We hypothesize that the work in the referenced project was only achievable in this form as research conducted within academia and that a similar effort may be less likely to occur in industry. 
Academia provides several conditions that are highly conducive to work on challenging projects such as this one \cite{digital_michelangelo}, which do not directly translate into financial returns in the future. On the other hand, large-scale research projects in industry may also have a similar impact if financial rewards can be expected upon their successful completion.

From the authors' perspective, large endeavors such as the Michelangelo project \cite{digital_michelangelo}, have the potential to move the field of computer vision forward, as well as to motivate researchers from diverse fields to work on problems that may contribute to future breakthroughs. 
We pose the hypothesis that the productive struggle exemplified by the "digital Michelangelo project" can be described by the theory of "Productive Failure" \cite{productive_failure}. 
We believe that similar large scale projects may, in the same way, lead to leaps in what will be possible in the future. To this end, we briefly discuss three topics which have the potential of making a profound impact:

\begin{itemize}
\item \textbf{Autonomous driving:} Perception for self-driving cars is naturally associated with intriguing challenges within the context of computer vision. Reliable, accurate scene "understanding" with useful representations for action are of utmost importance in this context.
\item \textbf{Virtual and augmented reality:}
Creating computer generated experiences is associated with other types of research opportunities. 
Virtual and augmented reality environments require a very realistic or appealing visualization mechanism. Likewise, interactivity in such settings will be a key component in the future. A moon-shot project proposed in this area is the idea of "digital human teleportation" with early results shown in \cite{photorealistic_li}.

\item \textbf{Bridging the gap between vision and other data modalities:}
In the recent past, such techniques have migrated to other disciplines and problem settings. A recent well-known case for successful interaction between different modalities is the Visual Question Answering task \cite{VQA} which combines textual and visual information. 
\end{itemize}

Again, all three of the mentioned aspiring goals for the field share the characteristic of being practical and being a stepping stone to several unsolved research questions. 

\begin{figure}
\includegraphics[width=.97\linewidth]{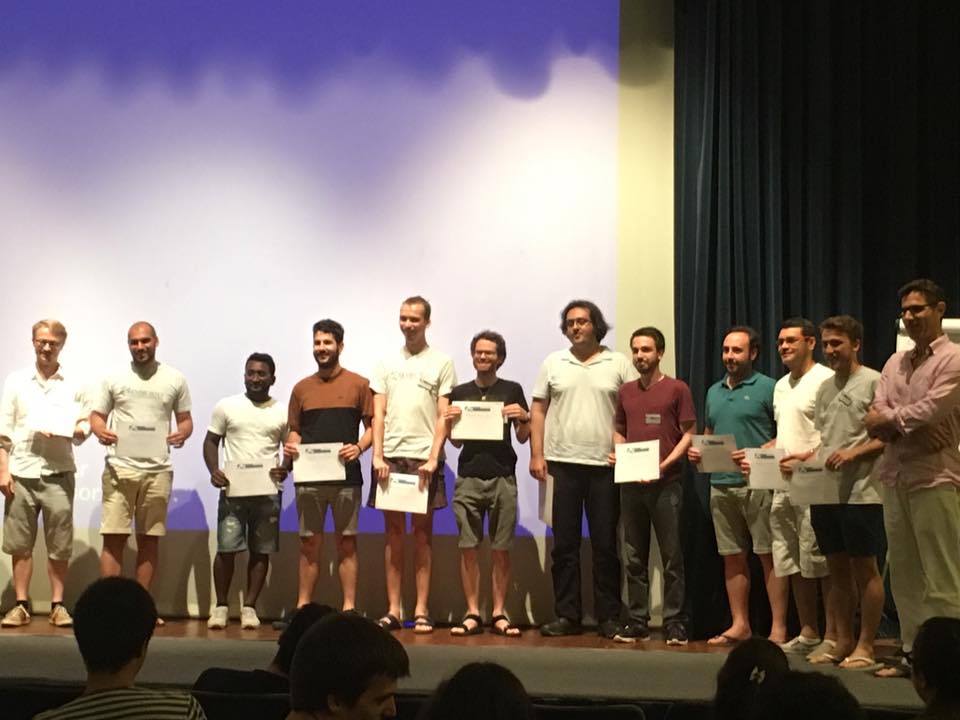}
\caption{Award ceremony of reading group participants with the organizer and creator of the reading group competition, Professor Stefano Soatto (right).}
\label{fig:reading_group}
\end{figure}

\section{Conclusion}

Comprehensive, interdisciplinary field research has pushed the envelope of what is possible in computer vision and improved the state of the art for many problems. The ramifications of such projects not only open new possibilities of research, but often also bring new technologies to society.

The "digital Michelangelo project" presented a richly detailed effort towards a solution to the problem of digitalization under certain constraints. It had a large impact on many fields of research, and important lessons can be distilled from it and it may be regarded as an archetype of productive struggle in science.

\textbf{Acknowledgements:} 
We are very grateful to Stefano Soatto who initiated the organization of the group work of which this report is a result. 

\bibliographystyle{icml2017}
\bibliography{main}

\end{document}